\documentclass[conference]{IEEEtran}
\IEEEoverridecommandlockouts
\usepackage{cite}
\usepackage{amsmath,amssymb,amsfonts}
\usepackage{algorithmic}
\usepackage{graphicx}
\usepackage{textcomp}
\usepackage{xcolor}
\usepackage{epstopdf}
\usepackage{bm}
\usepackage{subfigure}
\usepackage[hidelinks]{hyperref}
\usepackage{multirow}
\def\BibTeX{{\rm B\kern-.05em{\sc i\kern-.025em b}\kern-.08em
    T\kern-.1667em\lower.7ex\hbox{E}\kern-.125emX}}
\begin{document}

\title{Multi-typed Objects Multi-view Multi-instance Multi-label Learning}
%
\author{Yuanlin Yang$^{1,2}$, Guoxian Yu$^{1,2,3,4,*}$\thanks{$^*$Corresponding author: gxyu@sdu.edu.cn (Guoxian Yu). This work is supported by NSFC (No. 61872300, 62031003 and 62072380).},  Jun Wang$^3$, Carlotta Domeniconi$^5$,  Xiangliang Zhang$^4$\\
$^1$College of Computer and Information Sciences, Southwest University, Chongqing, China\\
$^2$School of Software, Shandong University, Jinan, China\\
$^3$Joint SDU-NTU Centre for Artificial Intelligence Research, Shandong University, Jinan, China\\
$^4$CEMSE, King Abdullah University of Science and Technology, Thuwal, SA \\
$^5$Department of Computer Science, George Mason University, VA, USA\\
Email: ylyang@swu.edu.cn; \{gxyu, kingjun\}@sdu.edu.cn, carlotta@cs.gmu.edu; xiangliang.zhang@kaust.edu.sa\\
}

\maketitle

\begin{abstract}
Multi-typed objects Multi-view Multi-instance Multi-label Learning (M4L) deals with interconnected multi-typed objects (or bags) that are made of diverse instances, represented with heterogeneous feature views and annotated with a set of non-exclusive but semantically related labels. M4L is more general and powerful than the typical Multi-view Multi-instance Multi-label Learning (M3L), which only accommodates single-typed bags and lacks the power to jointly model the naturally interconnected \emph{multi-typed} objects in the physical world. To combat with this novel and challenging learning task, we develop a joint matrix factorization based solution (M4L-JMF). Particularly, M4L-JMF firstly encodes the diverse attributes and multiple inter(intra)-associations among multi-typed bags into respective data matrices, and then jointly factorizes these matrices into low-rank ones to explore the composite latent representation of each bag and its instances (if any). In addition, it incorporates a dispatch and aggregation term to distribute the labels of bags to individual instances and reversely aggregate the labels of instances to their affiliated bags in a coherent manner. Experimental results on benchmark datasets show that M4L-JMF achieves significantly better results than simple adaptions of existing M3L solutions on this novel problem.
\end{abstract}


\section{Introduction}
With the prosperity of Internet of Things, objects are often represented by multiple heterogeneous feature views. For example, an image is numerically encoded by its texture, shape and color features. This image can also be simultaneously tagged with several related semantic labels (i.e., sun, sea, water, bird). To learn from such multi-modal multi-label data, various multi-view multi-label learning approaches have been introduced \cite{zhao2017mvlsurvey}. However, a real-world object may contain variable number of inconsistent instances (sub-objects). For example, a web page includes multiple images and content paragraphs, each of which can be viewed as an instance of the image/text view. To model such complex objects, Multi-view Multi-instance Multi-label Learning (\textbf{M3L}) has been invented, it aims to leverage the relationships between instances, their hosting objects (bags), semantic labels and between heterogeneous feature views to predict the labels of objects and those of individual instances \cite{nguyen2013M3LDA,yang2018M3DN,nguyen2014M3LMix,xing2019M3Lcmf,yang2020M3DNS}.

Existing M3L algorithms focus on \emph{single-typed} objects. Given the diverse interconnections between objects of multiple types, the labels of a complex object are not only determined by its own attributes, but also by its connections with objects of other types. M3L lacks the capability to simultaneously model \emph{multi-typed} objects. One typical solution is to project the objects of other types toward the target-type of objects to form the composite features, and then learn on the composite features (networks) \cite{gligorijevic2015DataFusion,yu2017brwlda}. Unfortunately, such projection may override the intrinsic structure information among multi-typed objects \cite{gligorijevic2015DataFusion,Zitnik2015DFMF}. Matrix factorization based solutions have been introduced to model interconnected multi-typed objects, and these solutions can respect the intrinsic structure of these objects and integrate multiple feature views of objects \cite{fu2018MFLDA,wang2020WMFLDA,Zitnik2019IFSurvey}. However, these solutions ignore the general case that one object is composed with multiple instances, which convey important context information for labeling the complex object \cite{Li2017MVMIL,xing2019M3Lcmf}.

\begin{figure}[ht]
\centering   
\includegraphics[width=8cm, height=7cm]{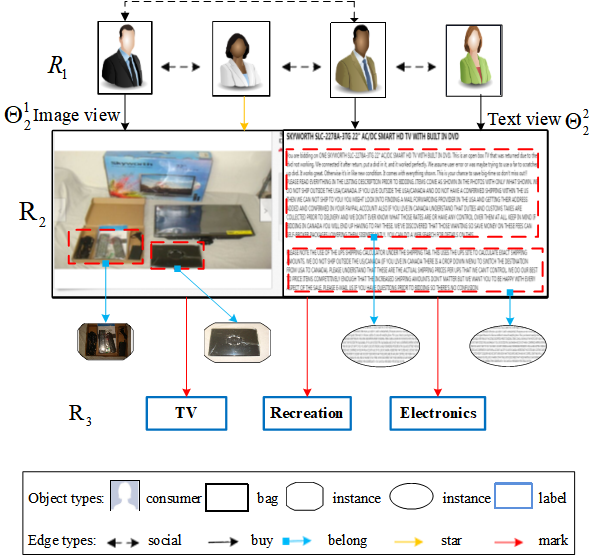}
\caption{An illustration of the M4 (Multi-typed objects Multi-view Multi-instance Multi-label) data with two types of objects (customers $R_1$ and goods $R_2$). The goods are encoded by the image view $\Theta_2^1$ and text view $\Theta_2^2$, and the goods (bags) in each view is further made of diverse instances (i.e., remote control panel, set-top box and prices), multi-type objects and their affiliated instances can be simultaneously tagged with several semantic labels $R_3$ (i.e., `recreation', `TV', `electronics'). M4L aims to fuse multi-type objects to annotate  the objects (instances)  with the semantic labels.}
\label{fig1}
\end{figure}

For example, as instantiated in Figure \ref{fig1}, a social network includes  user objects (encoded with social connections, personal profiles), image objects, text objects, and various instances (i.e., paragraphs and image parts) affiliated with these multi-typed objects. To effectively learn from multi-view multi-typed complex objects with interconnections, we term a new learning paradigm \emph{Multi-typed objects Multi-view Multi-instance Multi-label Learning} (\textbf{M4L}) and introduce a joint matrix factorization based solution to solve this challenging task. M4L takes the objects (bags), instances and labels as nodes, and constructs a heterogeneous network with diverse edge types to encode the inter- and intra-relations between multi-typed objects, associations between instances and their hosting bags, relations between bags and labels, and correlations between labels. Next, it jointly factorizes the block association data matrices of the heterogeneous network and the attribute data matrix of nodes of respective types into low-rank matrices to explore the latent representations of bags, instances and semantic labels, and then exploits the latent representations to predict the relations between bags (instances) and labels. In addition, to account for the bag-instance associations, we further introduce a dispatch and aggregation term to push the bag-level labels to individual instances and aggregate the instance-level labels to their hosting bags in a coherent way.

The main contributions of this work are:
\begin{itemize}
\item We study a \emph{new} learning paradigm Multi-typed objects Multi-view Multi-instance Multi-label Learning (M4L), which is a universal framework learning on naturally interconnected multi-typed complex objects. This learning scenario is more general than canonically studied M3L, which can only work with single-typed objects.
\item  We introduce a joint matrix factorization based solution M4L-JMF, which can leverage multiple inter(intra)-relations between bags of different types, associations between bags and instances, and the label correlations to annotate the complex bag and its instances.
\item  Experimental results on benchmark datasets show that M4L-JMF outperforms competitive and related methods.
\end{itemize}

\section{Related works}

Our work is closely related with M3L and its degenerated versions (Multi-view Multi-label Learning \cite{tan2018iMVML}, Multi-view Multi-instance Learning \cite{Li2017MVMIL}, and Multi-instance Multi-label Learning \cite{Zhou2012MIML}), and data fusion by matrix factorization \cite{Zitnik2015DFMF}. A comprehensive overview of the progress in these fast-progress areas is out of scope of this paper. Compared with these degenerated versions, M3L is less explored, due to the multiplicity and difficulty of learning from M3 data. To be self-inclusive, we give a brief review of M3L solutions.
To the authors knowledge, M3LDA \cite{nguyen2013M3LDA} is the first M3L algorithm, it learns
a visual-label part from the visual view and
a text-label part from the text view, and forces these two parts having consistent labels. M3DN \cite{yang2018M3DN} separately applies a deep network for each view, and requires the bag-level predictions from different views being consistent within the same bag. M3DNS \cite{yang2020M3DNS} extends M3DN by additionally making use of unlabeled instances and label correlations. M3Lcmf \cite{xing2019M3Lcmf} utilizes a heterogeneous network to capture different types of relations between bags, instances and labels, and then collaboratively factorizes the relational data matrices of the network into low-rank ones to explore the latent relationships between bags, instances, and labels. {WSM3L \cite{xing2020WSM3L} studies M3L in a more general setting with unpaired view data and missing labels by multi-modal dictionary learning.} However, these M3L methods \emph{can only} consider single-typed bags, while in practice, these bags are also connected with objects of other types (as shown in Figure \ref{fig1}), which indirectly reflect the property (i.e., labels) of the target bags.

Given the huge demand of integrating multi-modal data, data fusion techniques have been extensively studied and applied in various domains  \cite{gligorijevic2015DataFusion,Zitnik2019IFSurvey}. Compared with other data fusion solutions, such as feature view concatenation \cite{zheng2020feature} and classifier ensemble \cite{yu2012KDD}, matrix factorization based solutions can respect the intrinsic structure among multi-typed objects without projecting toward the target objects, and train a single model to simultaneously fuse multiple information sources with less information loss \cite{Zitnik2015DFMF}. To name a few, Wang \textit{et al.} \cite{wang2011SNMTF} applied symmetric nonnegative matrix tri-factorization
(SNMTF) to simultaneously cluster multi-typed objects. However, SNMTF has an overwhelming computation load, because it performs matrix factorization on a big matrix, whose block matrices encoded inter(intra)-relations between multi-typed objects. 
Data fusion by matrix factorization (DFMF) \cite{Zitnik2015DFMF} collaboratively factorized these block matrices with much smaller sizes into low-rank ones and then reconstructed the target relational matrix to predict the relations between multi-typed objects. \cite{fu2018MFLDA,wang2020WMFLDA} further considered the different relevance of multiple inter(intra)-relational block matrices toward the target prediction task and selectively fused these block data sources. All these prior studies of data fusion based on matrix factorization  simply assume that each object is made of a single instance. As such, they \emph{cannot} model the complex objects composed with diverse instances, as studied in M3L methods. {M4L is different from the heterogeneous information network based data fusion \cite{yu2020AHNF}. The latter focuses on the heterogeneity of objects and does not consider the composition (sub-objects) of complex objects, as the consumer-product relationship shown in Figure \ref{fig1}.} It simply takes consumers and commodities as nodes. Therefore, M4L considers a more sophisticated and practical learning scenario.


\section{The Proposed Method}
 \subsection{Problem Statement}
Suppose there are $m$ types of directly or indirectly related objects (including semantic labels and sub-objects, a.k.a. instances), which are encoded by a set of inter-relational data matrices $\mathbf{R}_{ij} \in \mathbb{R}^{n_i\times n_j}, i,j \in \{1,2,\cdots,m\}$.
One matrix $\mathbf{R}_{ij}$ encodes the relations between $n_i$ objects of the $i$-th type and $n_j$ objects of the $j$-th type,  and thus can be asymmetric. There are also a set of intra-association data matrices $\mathbf{\Theta}_i^{(t)}\in \mathbb{R}^{n_i \times n_i}$, $t \in \{1,2,\cdots, t_i\}$, where $t_i$ is the number of intra-relational data views for the $i$-th type of objects. Among these multi-typed objects (bags), some types of objects are further made of sub-objects (instances). Without loss of generality, we assume the $i$-th type of objects are instances of the $b$-th type, and $\mathbf{R}_{bi}(j,k)=1$ if the $k$-th object of the $i$-th type is a member instance of the $j$-th object of the $b$-th type (e.g., the remote control panel of the $4$-th type is an instance of the image object of the $2$nd type.)
Suppose each entity of the $m$-th type corresponds to a semantic label,  and the label correlations are encoded by the intra-relational data matrix $\bm{{\Theta}}_m \in \mathbb{R}^{n_m \times n_m}$.
When our target object type is $b$,
the aim of M4L is to predict the inter-relational matrix $\mathbf{R}_{bm}$ for $n_b$ bags and $n_m$ labels, and/or the inter-relational matrix $\mathbf{R}_{im}$ for $n_i$ instances and $n_m$ labels. The prediction is made by  learning a mapping function $f(\mathcal{R}, {\Theta})\in \{0,1\}^{n_m}$ to relate the objects to $n_m$ distinct labels.  Here, $\mathcal{R}$ collectively stores all the inter-relational data matrices $\mathbf{R}_{ij}$, and $\mathbf{\Theta}$ collectively stores all the intra-relational data matrices $\mathbf{\Theta}_i$.

\subsection{Joint Matrix Factorization}
To complete the relational data matrix $\mathbf{R}_{bm}$ (or $\mathbf{R}_{im}$) for bag (or instance)-label association prediction, we can take the target bags as anchors and then project objects of other types toward these anchors to form a composite bag-bag (instance-instance) intra-relational  data matrix, and then use the known labels of bags to predict the labels of other bags (or instances) of the same type. In fact, this projection idea has been extensively used to integrate interconnected multi-type objects, and worked with multiple kernel (view) learning, classifier ensemble based data fusion solutions \cite{zhao2017mvlsurvey,yu2013promk,yu2012KDD}. However, such projection may override the intrinsic structures among objects and cause information loss, and further compromise the performance \cite{Zitnik2019IFSurvey}.

Matrix factorization based data fusion techniques have been recently studied. They can integrate interconnected multi-typed objects of different types without conducting projection, while respect the intrinsic structures among objects \cite{Zitnik2015DFMF}. This basic framework can be formulated as follows:
\begin{equation}
\footnotesize
\begin{split}
\min_{\substack{\mathbf{G} \geq 0}} \Omega(\mathbf{G},\mathbf{S}) &=  \sum_{\substack{\mathbf{R}_{ij} \in \mathcal{R}}}{\|\mathbf{R}_{ij}-\mathbf{G}_i \mathbf{S}_{ij}\mathbf{G}_j^T }\|_F^2 \\ 
&+\sum_{t=1}^{\tau} tr(\mathbf{G}^T \mathbf{\Theta}^{(t)}\mathbf{G})
\end{split}
\label{eq1}
\end{equation}
The minimization of the above objective function aims to reconstruct the incomplete $\mathbf{R}_{bm}$ (or $\mathbf{R}_{im}$) to complete the associations between bags (instances) and labels, and thus achieve the prediction by integrating interconnected objects  of diverse types, without projecting these objects onto the target bags (instances).
Here, $\mathbf{G}=diag(\mathbf{G}_1, \mathbf{G}_2,\cdots,\mathbf{G}_m)$, $\mathbf{G}_i \in \mathbb{R}^{n_i\times k_i}$  is the low-rank representation of objects of the $i$-th type. $\mathbf{S}$ is made of $\mathbf{S}_{ij} \in \mathbb{R}^{k_i\times k_j}$ $(k_i{\ll}n_i,k_j{\ll} n_j)$, which can be viewed as a compressed data matrix that encodes latent inter-relations between objects of the $i$-th type and those of the $j$-th type.
Intra-association data matrices ${\mathbf{\Theta}}^{(t)}=diag({\mathbf{\Theta}}_1^{(t)},{\mathbf{\Theta}}_2^{(t)},\cdots,{\mathbf{\Theta}}_m^{(t)})$ $(t\in\{1,2,\cdots,max_it_i\})$, where the $i$-th block matrix along the main diagonal of ${\Theta}^{(t)}$ is zero if $t>t_i$, and $\tau=max_i t_i$ , ${\|\cdot\|}_F^2$ is the  Frobenius norm, $tr(\cdot)$ is the matrix trace operator.
Entries in intra-association data matrices are positive for dissimilar objects, and negative for similar ones. The positive entries can be viewed as \emph{cannot-link} constraints \cite{bilenko2004integrating}, which force pairs of dissimilar objects being far away from each other in the low-rank representation space.  These intra(inter)-association data matrices jointly guide the learning of mutually consistent low-rank matrix $\mathbf{G}_i$, since $\mathbf{G}_i$ is not only learnt by data matrices (i.e., $\mathbf{R}_{ij}$) directly related with the $i$-th type of objects, but also by data matrices (i.e., $\mathbf{R}_{jk}$, $j\neq i, k\neq i$) indirectly related with this object type, and by the intra-relational data matrices $\mathbf{\Theta}_i^{(t)}$.

A real-world object (bag) may be further made of several different sub-objects (instances), and the labels of this bag are determined by the labels of its instances \cite{Zhou2012MIML}. In many practical domains (i.e., medical image analysis and biology), the precise labels of instances are more important and interesting than those of bags, which carry more specific knowledge about the regions (i.e., local patches and  functional sites) of the bag (i.e., image and molecule) \cite{yu2020IsoFun}. Unfortunately, the labels of instances are typically unknown and while the labels of bags can be more easily collected. However, (\ref{eq1}) overlooks the important \emph{bag-instance} associations, and thus it can only model the degenerated fusion of multi-typed multi-view multi-label objects. Another disadvantage is that  (\ref{eq1}) does not differentiate the relevance of different relational data sources.

\subsection{Unified Objective Function}
To leverage the bag-instance association between two types of objects, we introduce a dispatch and aggregation term to push the labels of bags to their affiliated instances, and  reversely aggregate the labels of instances to their hosting bags at the same time. For this purpose, we extend  (\ref{eq1}) as follows:
\begin{equation}
\footnotesize
\begin{split}
&\min_{\substack{G \geq 0}} \Omega(\bm{G},\bm{S}) = \sum_{\substack{i,j=1}}^m{ \|\bm{R}_{ij}-\bm{G}_i\bm{S}_{ij}\bm{G}_j^T \|}_F^2\\
         &+\sum_{\substack{p=1}}^{m}\sum_{\substack{t=1}}^{\tau} tr(\bm{G}_p^T\bm{{\Theta}}_p^{(t)}\bm{G}_p)
                 +{\|\mathbf{R}_{bm}-\mathbf{R}_{bi}\mathbf{G}_i \mathbf{S}_{im} \mathbf{G}_m^T\|}_F^2
\end{split}
\label{eq2}
\end{equation}
$\|\mathbf{R}_{bm}-\mathbf{R}_{bi}\mathbf{G}_i \mathbf{S}_{im} \mathbf{G}_m^T \|_F^2$ is added to achieve the aggregation of the labels of instances (predicted by $\mathbf{G}_i \mathbf{S}_{im} \mathbf{G}_m^T$) to their hosting bags via the bag-instance association matrix $\mathbf{R}_{bi}$. Given the labels of bags are typically available, this term can also dispatch the labels of bags (stored in $\mathbf{R}_{bm}$) to individual instances. In this way, (\ref{eq2}) not only can integrate multi-typed objects, but also account for the complex objects that are made of multiple sub-objects (instances) to predict the labels of bags and those of instances in a coherent way.

Multiple inter(intra)-relational data matrices contain complementary information of objects of different types, but they may also include some noisy or irrelevant data matrices. Although the low-rank matrix factorization can reduce the inner noises of individual data matrices to some extent \cite{meng2013robust,chen2018cost}, it is still necessary to selectively fuse these relational data matrices with different relevance toward the target task. To concrete this, we advocate to set adaptive weights to intra(inter)-relational data matrices and thus to explicitly remove noisy data matrices as follows:
\begin{equation}
\footnotesize
\begin{split}
&\min_{\substack{G \geq 0}} \Omega(\bm{G},\bm{S},\bm{W}^r,\bm{W}^h) = \sum_{\substack{i,j=1}}^m{\bm{W}_{ij}^r\|\bm{R}_{ij}-\bm{G}_i\bm{S}_{ij}\bm{G}_j^T \|}_F^2\\
     &+\sum_{\substack{p=1}}^{m}\sum_{\substack{t=1}}^{\tau}\bm{W}_{pt}^htr(\bm{G}_p^T\bm{{\Theta}}_p^{(t)}\bm{G}_p)
     +{\|\mathbf{R}_{bm}-\mathbf{R}_{bi}\mathbf{G}_i \mathbf{S}_{im} \mathbf{G}_m^T\|}_F^2\\
     &+ \alpha \|vec(\bm{W}^r)\|_F^2 +\beta\|vec(\bm{W}^h)\|_F^2\\
    & s.t. \bm{W}^r{\geq}0, \bm{W}^h{\geq}0,\sum{vec(\bm{W}_{i}^r)}=1, \sum{vec(\bm{W}_{i}^h)}=1
\end{split}
\label{eq3}
\end{equation}
where $\mathbf{W}^r \in \mathbb{R}^{m \times m}$ and $\mathbf{W}^h \in \mathbb{R}^{m \times \tau}$  are the weight matrices. $\mathbf{W}^r$ stores the weights assigned to $|\mathcal{R}|$ inter-relational matrices and $\mathbf{W}_{pt}^h$ encodes the weight of the $t$-th intra-relational matrix of the $p$-th object type. $vec(\mathbf{W}^r_i)$ is the vectorisation operator that stacks the $i$-th row of $\mathbf{W}^r$. $\mathbf{W}^r_{ij}=0$ if $\mathbf{R}_{ij}\notin \mathcal{R}$. For $\bm{{\Theta}}_p^{(t)}$, if $t{\geq}max_it_i,\bm{W}_{pt}^h=0$. $\alpha$ and $\beta$ are the regularization weights for these two weight matrices. They work alike the ridge regression to avoid the trivial solution that selects only one inter-relational data matrix  and only one intra-relational data matrix. (\ref{eq3}) not only can explore the contribution of different intra-relational data matrices, but also selectively fuse inter-relational matrices by assigning weights to them.

{Suppose $T$ is the maximum number of iterations, the time complexity of our model is $O(T(|\mathcal{R}| + \tau m +m\tau n_m))$. $m$ is number of object types, $n_m$ represents the maximum number of objects of type $m$ and $\tau$  is the maximum number of views. The objective function of our M4L is non-convex in $\mathbf{G}$, $\mathbf{S}$, $\mathbf{W}^r$ and $\mathbf{W}^h$ altogether. We can use the idea of auxiliary functions frequently used in the convergence proof of approximate matrix factorization algorithms to alternatively optimize $\mathbf{G}$ and $\mathbf{S}$ in (\ref{eq3}) \cite{ding2008convex,lee2001nmf}.
 }



\section{Experiment}
\subsection{Experimental Setup}


{We   used two publicly available datasets (Isoform and LncRNA) for the  experiments, The   two  datasets have multi-typed objects and are specifically introduced below.

The Isoform dataset is collected from functional biology domain for predicting the functions of isoforms (instances), which are alternatively spliced from genes (bags) \cite{yu2020IsoFun}. This dataset is a natural testbed of M4L. Unfortunately, limited by the wet-lab techniques, it just has the bag-level labels (a.k.a. functional annotations of genes), so we use the bag-level predictions aggregated from instance-level for a surrogate evaluation. This evaluation protocol is canonically used in isoform function prediction \cite{luo2017functional,yu2020IsoFun}.  We randomly  selected 795 genes with 6,457 instances (isoforms), 495 miRNAs and 704 Gene Ontology terms (labels) as the dataset for experiments. The genes are represented with 2 feature views, and the isoforms are also represented with 2 feature views (Adrenal Gland and Esophagus Muscularis Mucosa). The more detailed information of these views can be found in \cite{yu2020IsoFun}.
The LncRNA dataset is also collected from biology domain, it contains six types of objects (LncRNAs(240), miRNAs(495), Genes(15527),  Drugs(8283), and Diseases(412)), it was widely used in predicting the association between lncRNAs and diseases \cite{fu2018MFLDA}. The genes are represented with 6 feature views, and the drugs with 10 feature views. The detailed information of multiple feature views  of this dataset can be found in \cite{fu2018MFLDA}.}

We perform experiments on the above benchmark datasets to quantitatively study the performance of the proposed M4L-JMF, and compare it against six representative and related approaches {(M3Lcmf\cite{xing2019M3Lcmf}, M2IL\cite{Li2017MVMIL} and ICM2L \cite{tan2020ICM2L}), and  data fusion solutions  (SNMTF \cite{wang2011SNMTF}, DFMF \cite{Zitnik2015DFMF} and MFLDA \cite{fu2018MFLDA}))}. The first three compared methods are M3L solutions or the degenerated versions, while the other three methods addressed the fusion of interconnected multi-typed objects via matrix factorization, without consideration of the bag-instance associations. The input parameters of these comparison methods are specified (or optimized) according to the recommendations of the authors in their code or papers. The sensitivity of these parameters will be studied later. For each compared method, we run 5-fold cross validation for 10 independent rounds, and report the average results.

To quantitatively evaluate the M4L model's performance, three widely used multi-label evaluation metrics (AUROC, AUPRC and AvgF1) are adopted. AUROC firstly computes the value of the area under the ROC curve for each label and then takes the average for all labels.   AUPRC firstly calculates the area under the Precision-Recall (PR) curve for each label and then takes the average values for all labels. {For these three evaluation metrics, their larger values indicate the better the performance. AvgF1 needs to convert the bag-label association probabilistic matrix   $\mathbf{R}_{bm}$ or instance-label matrix $\mathbf{R}_{im}$ into binary ones. Following the typical settings  \cite{yu2012KDD,fu2018MFLDA}, we adopt the top $K$ labels corresponding to the largest entries of  each row of $\mathbf{R}_{bm}$ ($\mathbf{R}_{im}$) as the relevant labels of bags (instances), where $K$ is the next integer of the average number of labels per bags/instances of the dataset.}

\begin{figure}[tp]
\centering
\includegraphics[width=6cm, height=4cm]{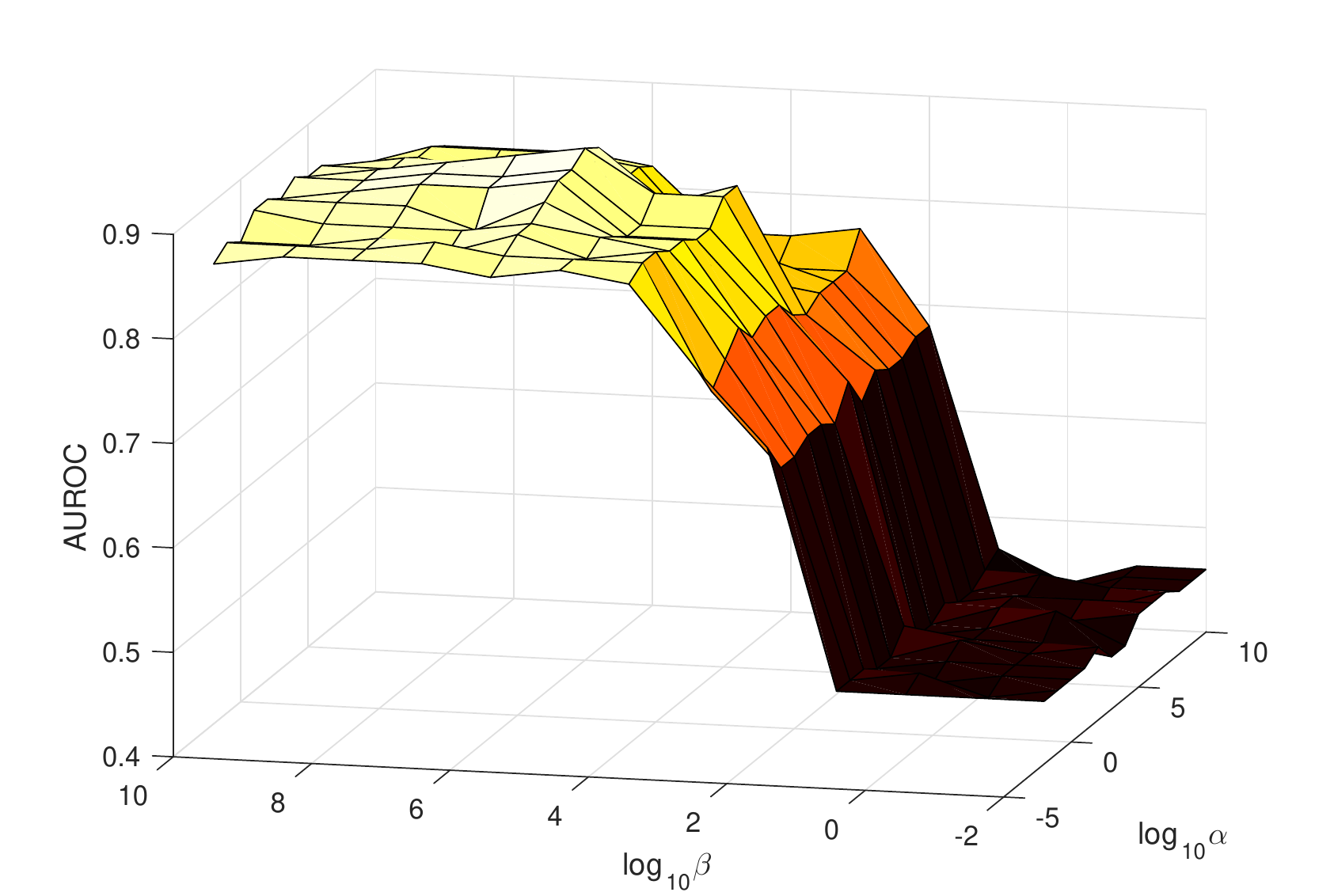}
\caption{AUROC of M4L-JMF under different settings of $\alpha$ and $\beta$ on the LncRNA dataset. }
 \label{fig3}
\end{figure}

\subsection{Results on M4 data}
To study the performance our M4L-JMF, we apply it on the Isoform dataset to predict the associations between isoforms and Gene Ontology terms, say the functional labels of isoforms. For comparison with M3L algorithms that are not designed to handle multi-type objects, we project objects of other types toward the genes at first and then form an M3 dataset composed with genes, isoforms and Gene Ontology labels, and then apply M3Lcmf, ICM2L, and M2IL on this projected M3 dataset. For matrix factorization based data fusion algorithms, we skip the bag-instance associations but fuse these types of objects to predict the association between labels and genes. Table \ref{tab3} summarizes the results of M4L and compared methods.

From Table \ref{tab3}, we have the following important observations:\\
\noindent(i) {Projection of multi-type objects along the target type causes information loss.
To handle multi-typed objects,
M3L methods (M3Lcmf, ICM2L and M2IL) were applied on the projection of multi-typed objects. The results in Table \ref{tab3} show that M3L methods in general perform worse than matrix factorization based methods and M4L-JMF, evaluated on AUROC and AUPRC.} This comparison corroborates the issue of fusing data by projecting data along the target type, which has been typically used in previous data fusion solutions \cite{yu2017brwlda,lu2018prediction,yu2013promk}. Our proposed M4L-JMF can better deal with multi-type objects than the simple adaption of M3L solutions.

\noindent(ii) The bag-instance associations carry important information  and can boost the performance of M4L. {This observation is made by the comparison between matrix factorization(MF)-based  solutions and M4L-JMF. Due to the ignorance of bag-instance associations, MF solutions perform worse than M4L-JMF on all metrics.}
In addition, M3Lcmf considers all the inter(intra)-relation between bags, instances and labels. Therefore,   it has the highest AvgF1. However, it has lower AUROC and AUPRC than MF solutions and M4L due to its limitation on handeling multi-type objects, as we discussed in the observation (i).
Overall, these comparisons confirm that it is important to model the bag-instance association.

\noindent(iii) Selective fusion of data can further boost the performance. M3Lcmf and DFMF simply add up all the inter(intra)-relational data matrices, without considering the different relevance of these data sources. In contrast, our proposed M4L-JMF differentiates the relevance of these data sources. For this reason, M4L-JMF manifests much better results than them. Although MFLDA also considers the different relevance of inter-relational data sources, they ignore the intra-relational data sources. So it also gives lower results than our proposed M4L-JMF.

In summary, the results on the real M4 dataset confirm that M4L-JMF can more comprehensively model M4 data, without projecting the multi-type objects and skipping the bag-instance associations. For this advantage, it achieves better results than these compared methods.


\begin{table}[tbp]

   \centering
   \scriptsize
   \caption{ Results on Isoform dataset of M4L, M3L-based methods and matrix factorization (MF)-based methods by 5-fold cross validation. $\bullet/\circ$ indicates whether M4L-JMF is statistically (according to pairwise $t$-test at $95\%$ significance level) superior/inferior to the other method.}
   \label{tab3}
		\begin{tabular}{l l l l l }\hline
& \textbf{Method}&                       \textbf{ AvgF1}&                  \textbf{AUROC}&                \textbf{AUPRC}\\ \hline
\multirow{3}{*}{M3L $\qquad$}&
M3Lcmf&                    0.152$\pm$0.004$\circ$&     0.663$\pm$0.018$\bullet$&    0.154$\pm$0.013$\bullet$\\
& ICM2L&                     0.074$\pm$0.001$\circ$&     0.533$\pm$0.001$\bullet$&    0.041$\pm$0.022$\bullet$\\
& M2IL&                      0.025$\pm$0.004$\bullet$&   0.544$\pm$0.009$\bullet$&    0.032$\pm$0.013$\bullet$\\
\hline
\multirow{3}{*}{MF $\qquad$}& DFMF&                      0.051$\pm$0.001$\bullet$&   0.943$\pm$0.009$\bullet$&     0.637$\pm$0.054$\bullet$\\
& SNMTF&                     0.021$\pm$0.001$\bullet$&   0.790$\pm$0.012$\bullet$&     0.015$\pm$0.002$\bullet$\\
& MFLDA&                     0.029$\pm$0.002$\bullet$&   0.946$\pm$0.005$\bullet$&     0.546$\pm$0.011$\bullet$\\
\hline
M4L $\qquad$ & M4L-JMF&                       0.055$\pm$0.002&             0.967$\pm$0.004&                  0.674$\pm$0.026\\
      \hline
		\end{tabular}
\end{table}

\subsection{Results on  LncRNA dataset}
We further study the performance of M4L-JMF on the LncRNA dataset (a natural test for multi-type objects fusion). For the experiments on the LncRNA dataset (without instance-label associations), we skips the bag-instance associations and compare M4L-JMF with matrix factorization based solutions only. The other pre-processes are the same as the experiments in previous subsection. The results on LncRNA dataset are given in Table \ref{tab5}.

From Table \ref{tab5}, we can find that, even without the important bag-instance associations, M4L-JMF still shows a good performance in fusing multi-type objects. That is because M4L-JMF can selectively fuse both the inter-relational and the intra-relational data sources, whereas these compared methods either ignore the different relevance of these data sources, or only differentiate the inter-relational ones.
\begin{table}[tbp]
   \centering
   \scriptsize
   \caption{ Results of M4L-JMF and matrix factorization based methods on the LncRNA dataset. $\bullet/\circ$ indicates whether M4L-JMF is statistically (according to pairwise t-test at $95\%$ significance level) superior/inferior to the other method.  }

   \label{tab5}
\begin{tabular}{c l  l l l}\hline
	\textbf{Method}&                     \textbf{AvgF1}&                  \textbf{AUROC}&                \textbf{AUPRC}\\ \hline
DFMF&                    0.062$\pm$0.001$\bullet$&   0.872$\pm$0.007$\bullet$&    0.546$\pm0$.091$\bullet$\\
SNMTF&                   0.023$\pm$0.001$\bullet$&   0.804$\pm$0.001$\bullet$&    0.016$\pm$0.002$\bullet$\\
MFLDA&                   0.064$\pm$0.003$\bullet$&   0.874$\pm$0.005$\bullet$&    0.573$\pm$0.053$\bullet$\\
M4L-JMF&                 0.067$\pm$0.002&          0.895$\pm$0.004&           0.616$\pm$0.026\\
      \hline
		\end{tabular}
\end{table}

Overall, these experimental results demonstrate the flexibility of M4L-JMF in diverse settings, and justify the contributions of weighting inter(intra)-relational data matrices.


\subsection{Parameter Sensitivity Analysis}


\begin{figure}[tp]
\centering
\subfigure[ low rank size ]{\includegraphics[height=4cm,width=4cm]{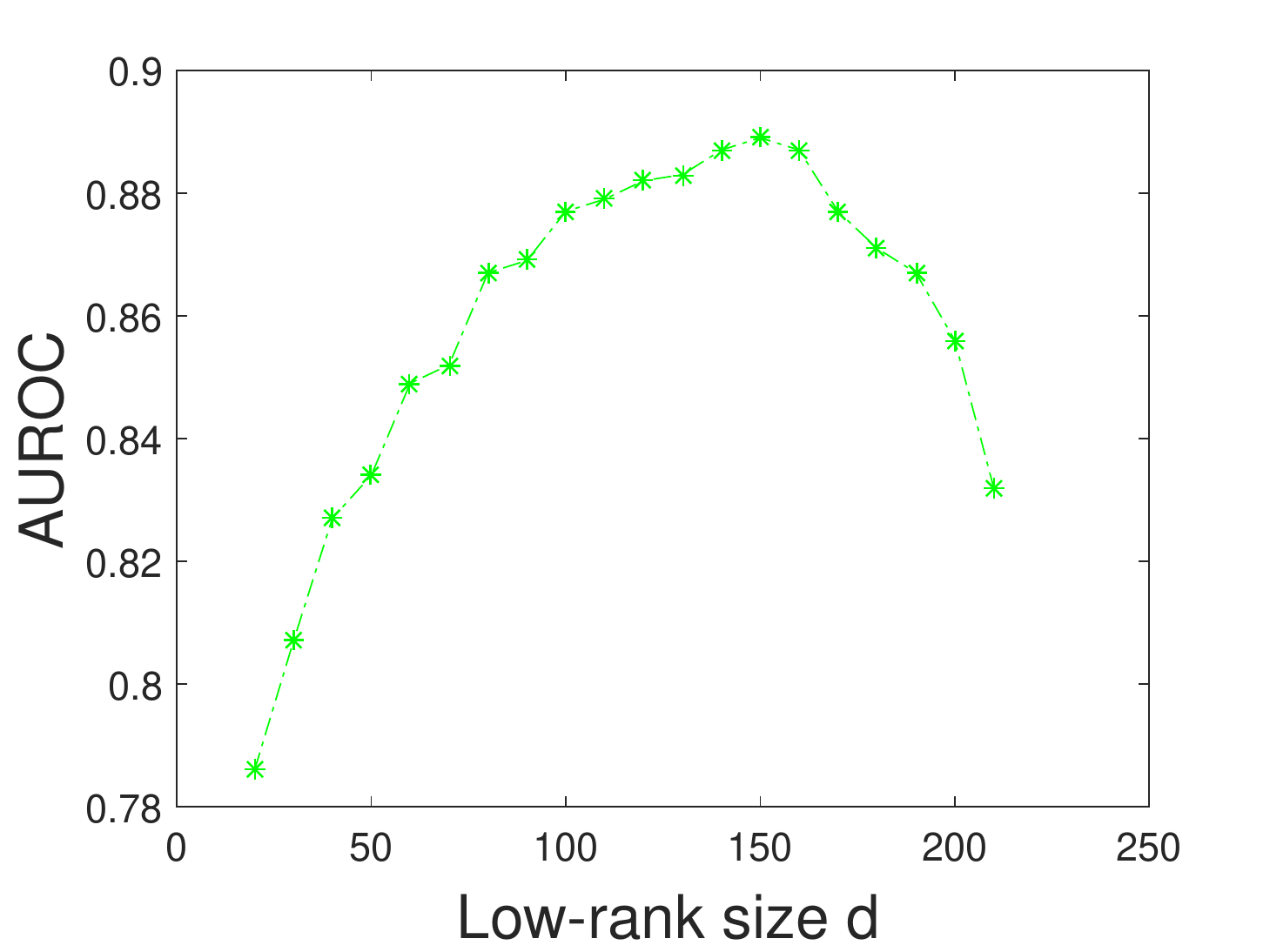}}\label{fig4a}
\subfigure[ Convergence trend ]{\includegraphics[height=4cm,width=4cm]{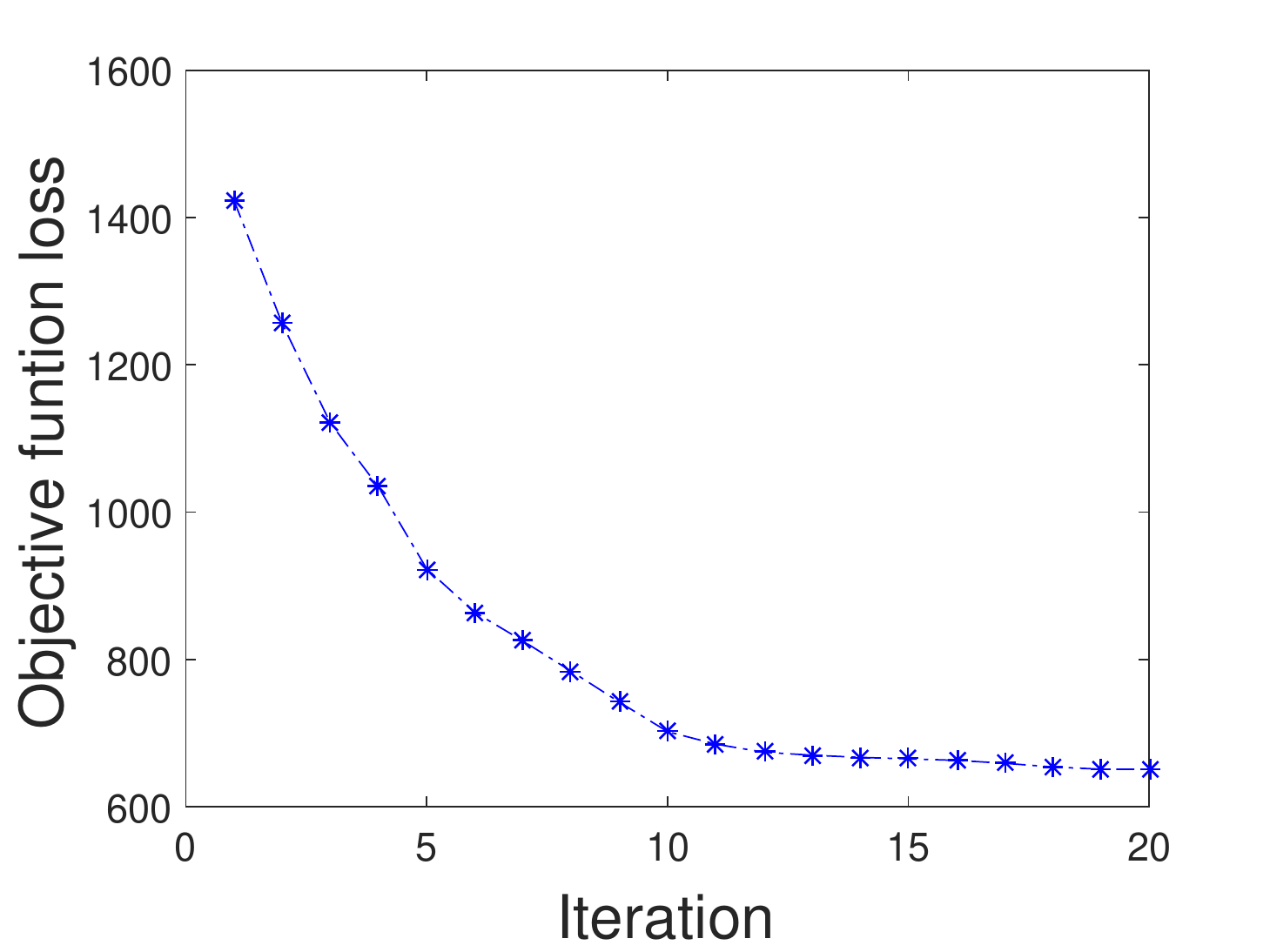}}\label{fig5a}

\caption{ AUROC vs. $d$ (low rank size of $\textbf{G}_i$) and convergence trend on the LncRNA dataset}
\label{fig4}
\end{figure}

%

{In this paper, three parameters ($\alpha$, $\beta$ and the low-rank size $d$ of $\textbf{G}$) in (\ref{eq3}) should be specified for our proposed M4L-JMF. To investigate the sensitivity of first two parameters, we vary $\alpha$ and $\beta$ in the range $\{10^{-2}, 10^{-1}, \cdots ,10^{10}\}$, and report the average AUROC of M4L-JMF under different combinations of them in Fig. \ref{fig3}. M4L-JMF achieves the highest AUROC when $\alpha=10^6$ and $\beta=10^7$.  The AUROC value increases when $\alpha$ or $\beta$ rises,  then it slightly decreases when $\alpha>10^6$ or $\beta>10^7$.  A too small input value for  $\alpha$ (or $\beta$) makes M4L-JMF only fuse one inter-relational data matrix and one intra-relational data matrix.  On the other hand, a too large value for $\alpha$ (or $\beta$) leads M4L-JMF fusing all the relational data matrices without differentiating the relevance among them. This pattern shows that M4L-JMF can mine the complementary information of multiple relational data sources and account for different relevance of them.} We also vary the low-rank size $d$ ($k_i=d$ for all object types for simplicity) for the representation of objects ($\textbf{G}_i$) from different ranges to study the optimal low-rank size using 5-fold cross validation, and show the AUROC under each input value of $d$ with $\alpha=10^6$ and $\beta=10^7$ in Fig. \ref{fig4}(a). For the LncRNA dataset, we observe that a too small $d$ can not sufficiently encode the latent feature information of multi-type objects and labels, and while a too large $d$ may bring in some noises and thus leads to a low AUROC value.
To investigate the convergence trend of M4L-JMF, we record the objective function value, {i.e., the loss of (\ref{eq3})  in each  iteration  on  LncRNA  datasets,} and report the results in Fig. \ref{fig4}(b). We can see that the loss decreases as the iteration proceeds and comes to a convergence within 20 iterations. This trend proves that our alternative optimization procedure can quickly converge.

\section{Conclusions}
In this paper, we studied a novel learning paradigm (Multi-typed objects Multi-view Multi-instance Multi-label Learning) for naturally interconnected multi-type complex objects, and introduced a joint matrix factorization based approach M4L-JMF. Experimental results on real-world and benchmark datasets validated that M4L-JMF can more comprehensively fuse multi-type objects and mine complex relations between  bags, instances and labels, and it achieves better results than other competitive and related methods.

\bibliographystyle{IEEEtran}
\bibliography{M4L_JMF}
\end{document}